\documentclass{article} 
\usepackage{iclr2024_conference,times}

\usepackage{amsmath,amsfonts,bm}

\def\eqref#1{equation~\ref{#1}}

\def\1{\bm{1}}

\def\vone{{\bm{1}}}

\DeclareMathAlphabet{\mathsfit}{\encodingdefault}{\sfdefault}{m}{sl}
\SetMathAlphabet{\mathsfit}{bold}{\encodingdefault}{\sfdefault}{bx}{n}

\usepackage{hyperref}
\usepackage{url}
\usepackage{tcolorbox}
\usepackage{booktabs}
\tcbuselibrary{skins}
\definecolor{NavyBlue}{RGB}{0,0,128}
\definecolor{ForestGreen}{RGB}{34,139,34}
\definecolor{Gray}{HTML}{808080}
\usepackage{listings}
\lstdefinestyle{mystyle}{
    keywordstyle=\color{magenta},
    basicstyle=\ttfamily\tiny,
    breakatwhitespace=false,         
    breaklines=true,                 
    keepspaces=true,                 
    showspaces=false,                
    showstringspaces=false,
    showtabs=false,                  
    tabsize=1,
    numbers=none
}
\title{Thinking LLMs: 
General Instruction \\Following with Thought Generation}

\renewcommand\sup[1]{$^{#1}$}
\author{Tianhao Wu\sup{1,2} \quad Janice Lan\sup{1} \quad ~~~~~Weizhe Yuan\sup{1,3} \\ \textbf{Jiantao Jiao}\sup{2}\quad \textbf{~~Jason Weston}\sup{1,3}  \quad \textbf{Sainbayar Sukhbaatar}\sup{1}
\\ \\
\textsuperscript{1}Meta FAIR \quad
\textsuperscript{2}University of California, Berkeley \quad
\textsuperscript{3}New York University
}

\newcommand{\method}{Thought Preference Optimization}
\newcommand{\methodshort}{TPO}
\newcommand{\eft}{STE}

\usepackage{tablefootnote}
\usepackage{enumitem}
\usepackage{array}

\newtcolorbox{prompt}[1]{
    enhanced,
    drop shadow=black!5!white,
    left=2mm,
    right=2mm,
    top=2mm,
    bottom=2mm,
    boxsep=0mm,
    rounded corners,
    title=#1,
    colframe=black!50!white,
    fontupper=\footnotesize\linespread{0.9}\fontfamily{lmr}\selectfont,
    fontlower=\footnotesize\linespread{0.9}\fontfamily{lmr}\selectfont,
    }

\iclrfinalcopy 
\begin{document}

\maketitle

\begin{abstract}
LLMs are typically trained to answer user questions or follow instructions similarly to how human experts respond. However, in the standard alignment framework they lack the basic ability of explicit thinking before answering. Thinking  is important for complex questions that require reasoning and planning -- but can be applied to {\em any} task. We propose a training method for equipping existing LLMs with such thinking abilities for general instruction following without use of additional human data.
We achieve this by an iterative search and optimization procedure that explores the space of possible thought generations, allowing the model to learn how to think without direct supervision.
For each instruction, the thought
candidates are scored using a judge model to evaluate their responses only, and  
then optimized via preference optimization. 
We show that this procedure leads to superior performance on AlpacaEval and Arena-Hard, and  shows gains from thinking on  non-reasoning categories such as
marketing, health and general knowledge, in addition to more traditional reasoning \& problem-solving tasks.  
\end{abstract}
\section{Introduction}

Large Language Models (LLMs) are based on the Transformer architecture \citep{vaswani2017attention} that predicts the next token at each step.
Each token takes the same amount of compute, so 
when LLMs are prompted with a user instruction, they have a fixed compute budget to generate the first response token regardless of the instruction's complexity.
One way to increase the compute budget for harder instructions is to allow LLMs to \emph{think} internally before outputting an response.
This is similar to humans who will take more time and think before answering complex questions. 
%
%

One approach is to generate thoughts as text, 
which  takes advantage of the natural language capabilities of LLMs. LLMs   are pre-trained on text containing human-written thoughts, which are hence encoded into the model.
Chain-of-Thought (CoT) \citep{wei2022chain} is a widely used prompting technique that elicits such behavior by
asking the model to write down its reasoning steps.
However, the usage of CoT has been mostly limited to math and reasoning tasks.
Meta-analysis by \citet{sprague2024cot} found CoT methods to be unhelpful on tasks that do not involve math and logic.

In this paper, we focus on general instruction following instead of focusing on math or logic tasks.
We argue that ``thinking'' should have broad utility.
For example, in a creative writing task, internal thoughts can be used to plan overall structure and characters.
In other tasks, internal thoughts can be used for understanding the user instruction better.
Of course, it is likely that less thinking is required for simpler tasks, and more thinking for more complex ones. 
In general, we hypothesize that such \emph{Thinking LLMs} will have an advantage on all sufficiently complex tasks.
The emergence of recent commercial products like OpenAI-O1~\citep{openai_o1} also support our motivation.

However it is challenging to train a model to think due to the lack of supervised training data.
Although pre-training data does contain valuable information, coverage can be limited in certain domains as internal thoughts are often omitted in human writing.
Existing post-training datasets  typically consist of human responses, or preferences over responses, with no information on thought processes.
The same is true for existing reward models.
Combined with the difficulty and cost considerations of collecting human thought data, these factors impose a barrier in training Thinking LLMs.

The goal of this paper is to investigate the possibility of converting existing LLMs into Thinking LLMs that work across a wide variety of tasks, without any addition data.
To this end, we introduce \method{} (\methodshort{}) that further trains an instruction-tuned LLM to make it capable of having internal thoughts.
Our method is simple and reuses many parts of existing 
training pipelines.
The LLM is first instructed to produce an output sequence that can be divided into thought and response parts.
The thought part is considered internal, and not part of the response shown to the user. 
We optimize this thought and response output through iterative Reinforcement Learning from AI Feedback (RLAIF) training. 
We rely on a standard judge model that is trained to evaluate responses only, and implicitly judge the quality of the thoughts via the induced responses. 
This has the advantage of not requiring human curated thoughts or a special judge model capable of evaluating thoughts.
Through preference optimization the thoughts are then optimized to improve the resulting responses.
This contrasts with recent works such as \citet{snell2024scaling, kumar2024training} which use additional supervision signals to guide models in self-refinement or self-correction. Instead of directly guiding the internal thought process, we allow the model to independently \textit{learn to think}.

Given the multitude of evidence showing the effectiveness of CoT on logic-based tasks like math or coding, we focus our experiments on general instruction following instead.
We train on diverse user instructions and evaluate our models on AlpacaEval and Arena-Hard, benchmarks that test general instruction following.
We obtain a strong win rate of 52.5\% and 37.3\% respectively on them, outperforming the direct LLM counterpart without explicit thinking. 
We also conduct more fine-grained evaluations to determine which types of instructions benefit from thought.
Surprisingly, we observe that thinking not only benefits topics like reasoning and problem solving, but also leads to better performance on categories that are not typically considered in the reasoning domain,
such as general knowledge, marketing, and health.
This opens up a new opportunity to develop Thinking LLMs aimed at general instruction following rather than specializing in more narrow technical fields. 


\section{\method{}}
\begin{figure}
    \centering
    \includegraphics[width=1\linewidth]{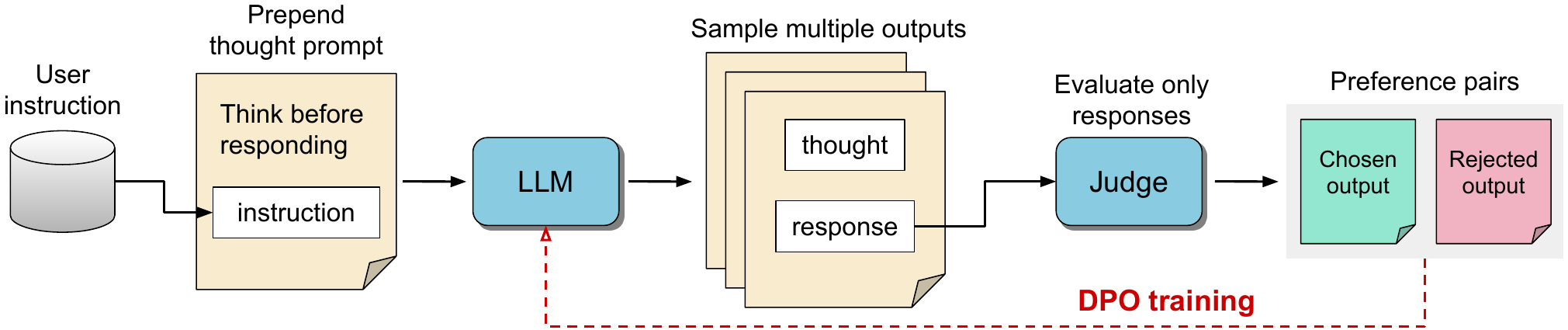}
    \caption{\textbf{\method{}:} We start by prompting the LLM to generate thoughts before its response. After sampling different outputs, we feed the response parts to the judge model which determines the best and worst ones. Then we use the corresponding full outputs as chosen and rejected pairs for DPO optimization. We perform multiple iterations of this training.}
    \label{fig:method}
\end{figure}

We now describe our \method{} (\methodshort{}) method for teaching LLMs to think before responding, as depicted in \autoref{fig:method}.
We start with a typical instruction-tuned LLM that outputs a response directly after the user instruction.
We  assume that there is no provided labeled thought data that we can finetune on, which makes training much more challenging.
Instead, as a starting point to bootstrap our training process, for a given training user instruction, we prompt the model to generate its thought process followed by the response.
Sampling multiple such outputs,  we then use preference optimization to improve the quality of thoughts (and paired responses) based solely on the quality of the responses.

\subsection{Generating Thoughts from Thinking LLMs}
\label{sec:generate_thought}

\begin{figure}[t]
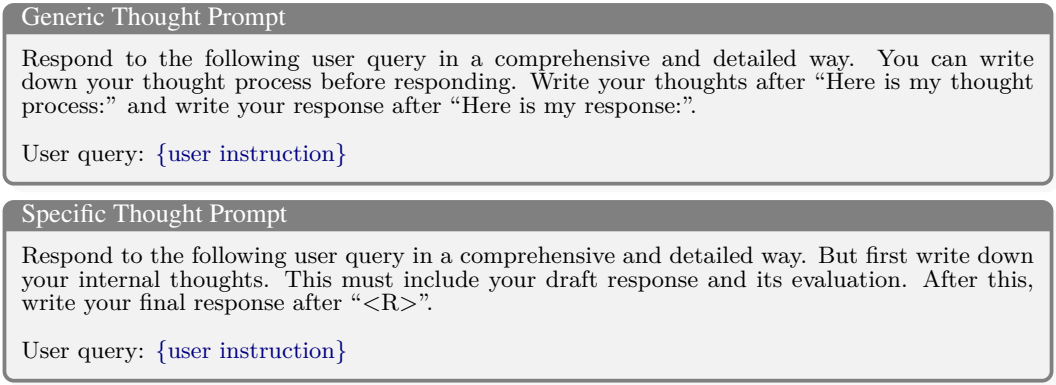

    \centering
    \begin{prompt}{Generic Thought Prompt}
    \label{prompt:meta}   
    Respond to the following user query in a comprehensive and detailed way. You can write down your thought process before responding. Write your thoughts after ``Here is my thought process:'' and write your response after ``Here is my response:''. \\
    \\
    User query: {\color{NavyBlue}
    \{user instruction\}
    }
    \end{prompt}
    \begin{prompt}{Specific Thought Prompt}
    Respond to the following user query in a comprehensive and detailed way. But first write down your internal thoughts. This must include your draft response and its evaluation. After this, write your final response after ``\textless R\textgreater''.
    \\
    \\
    User query: {\color{NavyBlue}
    \{user instruction\}
    }
    \end{prompt}
    
    \label{fig:prompt2}
    \caption{{\bf Thought Prompting}. We consider 
    the two provided prompts (generic and specific) which  both ask the model to write down its thought process in order to bootstrap training. 
    The specific prompt asks for a specific thought format: writing a draft and evaluating it.
    Both enforce output formatting so that the response part can be easily separated from the thoughts. 
     \label{prompt:std}
    }
    \label{fig:prompt}
\end{figure}

Ideally, thought generation should be simple and compatible with existing LLM infrastructures.
Hence, we keep the model architecture the same, as an autoregressive Transformer, although our method is potentially compatible with any model that outputs a sequence of tokens.
At inference time, the core process is that the output consists of two parts:  a thought part followed by a response part, both of which are in natural language.
After generation, instead of directly sending that entire token sequence to the user,  we preprocess it by splitting it into the two parts, and only sending the response part.

At the beginning of our training, we achieve this by prompting the model to write its thought process.
We consider two possible \emph{thought prompts},  shown in \autoref{fig:prompt}.
In order to separate thought from answer, we need the model to follow a strict format.
The thought prompts contain fixed keywords the model should use so that we can use simple string matching to locate where the response part begins.

\paragraph{Thought Prompt Types}
While the training process will change and optimize the type of the thoughts, the initial thoughts are still important as they act as a starting point.
The first thought prompt given in \autoref{fig:prompt} (top) is more generic and leaves it up to the model what the thoughts will contain.
We also experiment with a more \emph{specific} thought prompt, given in \autoref{fig:prompt} (bottom),
that specifies that the thought should contain a draft response and its evaluation.
Such specific prompts give us more control over the content of the thoughts, but also requires expert knowledge about what type of thoughts are helpful in LLMs.

\paragraph{Making Thoughts Internal}
As we mentioned, the thought part will be hidden from the end user, and only the response part will be provided to them.
This differentiates our outputs from CoT prompting where the reasoning steps typically become part of the overall response, sometimes without there being a clear distinction.
While the latter might be useful in certain cases like solving math problems, in general the user expects to receive a response without excessive intermediate reasoning steps.
Hiding the thought part allows it to take many forms that are usually not interesting to the user: making mistakes, drafting responses and evaluating them, trying to understand the question better, etc.
Of course, we can also give an option to reveal the thought part to the user for the purpose of interpretability and for analysing the underlying thought process behind the response.

In theory, the thoughts can take any form 
that is comprised of generated tokens, 
and do not even have to be in natural language.
Their primary goal is to allow the model to perform extra computation to improve the quality of the response \citep{pfau2024let}.
However, thoughts in natural language have several benefits such as taking advantage of human-written LLM pre-training data, and  
allowing humans to inspect and interpret the behaviour of the model.
Hence, we use this setting because current LLMs can generate thoughts well in natural language.

\subsection{Optimizing Thoughts via Preference Optimization}
While our initial thought prompting generates thoughts via the instruction tuned model, they are not optimized to be actually useful in making the response better. We  find they typically underperform {\em thoughtless} direct responses, which instruction-tuned LLMs have been heavily optimized for.
Therefore, we need to train our model so it makes better use of thought generation.
We employ the Reinforcement Learning from AI Feedback (RLAIF) paradigm \citep{bai2022constitutional,zhu2024starling} where we generate from the model and rank its responses using a reward model that acts as a judge.
In particular, we use iterative Direct Preference Optimization (DPO) \citep{rafailov2024direct,xu2023some} for its simplicity and efficacy.

Unlike conventional RLAIF, we will not feed the whole model output to the judge.
Instead, the judge can only see the response part of the outputs, so the thought part cannot influence its judgement.
We chose this approach for several reasons.
First, there is a lack of a judge model that is capable of evaluating internal thoughts.
Building such a judge is inherently challenging because it is hard to collect human thoughts.
In any case, even if such data was collected, it is not clear if human-written thoughts will be equally useful for LLMs.
Secondly, the ultimate goal is to provide better responses to the user.
Thus, it might be better to optimize the final objective instead of relying on an auxiliary objective that might not align well.

Our training starts with a seed model $\mathcal{M}_0$ that is instruction-tuned to directly respond to the user instruction.
We also need a dataset of user instructions $\{x_i\}$ to begin training the model.
At each training iteration $t$, we feed instructions to the current model $\mathcal{M}_t$ along with our thought prompt $p$ as described in \autoref{sec:generate_thought}:
\[
\mathcal{M}_t(p + x_i) \rightarrow \{z_i^k, y_i^k\}.
\]
Here ``+'' means the prompts are concatenated as input context to the LLM.
For each input, we sample $k \le K$ outputs, each containing thought $z_i^k$ and response $y_i^k$ parts.

\paragraph{Building Preference Pairs}
After extracting the response parts $y_i^k$, we feed them to the judge model $\mathcal{J}$ for scoring.
For pointwise judge models that take a single response and output a scalar score, the process is simple:
\[
\mathcal{J}(x_i, y_i^k) \rightarrow s_i^k \in \mathbb{R}.
\]
We also consider  judge models that take a pair of responses and output the winner. In this case we apply the judge model to all possible pairs $\{y_i^m, y_i^n\}$  from the set of responses.
This includes swapping positions in order to reduce the position-bias of the judge.
Once we have all pairwise winners, we convert those to individual pointwise scores $s_i^k$ using ELO scoring as performed in \cite{Wu2024MetaRewardingLM}. See \autoref{elo_computation} for more details.


Next, we select the highest and lowest scoring responses as ``chosen'' and ``rejected'' samples to construct a preference pair.
Note that the preference pairs contain both thought and response parts.
\[
\text{Pair} = \{p + x_i \rightarrow z_i^c + y_i^c\ ;\ p + x_i \rightarrow z_i^r + y_i^r\} \quad \text{where} \quad c = \text{argmax}_k s_i^k \quad r = \text{argmin}_k s_i^k.
\]
Using this process,  the model can learn which thought led to a better response.

\paragraph{Iterative Training} Once we have built preference pairs, we use them with the DPO loss to train the current model $\mathcal{M}_{t}$. 
This gives us a new model $\mathcal{M}_{t+1}$ that will be used for the next training iteration.
Note that we do not use data derived from previous iterations for training the current iteration, under the assumption that they are lower quality.
In addition to DPO, we also experiment with the IRPO loss \citep{pang2024iterative} that combines DPO with the NLL loss.

\paragraph{Length-Control}
It is known that some judge models tend to favor longer responses \citep{dubois2024length,yuan2024following}. This length-bias causes the response length to grow with each training iteration, resulting in an overly verbose model.
To mitigate this, we implement a length-control (LC) mechanism. 
Let us define a normalization function $N(l_i^k) = \left(l_i^k-\operatorname{mean}_k(l_i^k)\right)/\operatorname{std}_k(l_i^k)$.
We recompute the scores by penalizing longer responses
$$s_i^k \leftarrow N(s_i^k) - \rho N(l_i^k).$$
The hyper-parameter $\rho$ controls the strength of the length-control mechanism.
Note we normalize both the score and the length to align them into a similar scale.



\section{Experiments}
\label{sec:results}
\subsection{Setup}
We use Llama-3-8B-Instruct \citep{dubey2024llama} as a seed model in our training.
As a judge model, we consider two choices of model: Self-Taught Evaluator (\eft{}) \citep{Wang2024SelfTaughtE} and ArmoRM \citep{wang2024interpretable}.
\eft{} is a LLM-as-a-Judge model based on Llama-3-70B-Instruct.
Given two responses, it outputs its preference in natural language after generating a CoT.
ArmoRM is a 8B reward model that directly outputs a scalar score to a single response.

For initial experiments, we use the synthetic instructions from \cite{yuan2024selfrewarding} for training.
These instructions are generated from Llama-2-70B-Chat using 8-shot prompting consisting of random samples from the Open Assistant dataset \citep{kopf2024openassistant}. 
For later experiments, we switched to UltraFeedback \citep{cui2023ultrafeedback}, which contains actual human instructions.
Each training iteration uses 5000 instructions that were not part of the previous iterations.

We generate $K=8$ responses per prompt using temperature 0.8 and top-p of 0.95.
We train for 10 epochs in each iteration and select the best checkpoint using a validation set of 1500 prompts randomly sampled from UltraFeedback.
We perform up to 4 iterations.
We usually set the length-control parameter $\rho\in [0,0.5]$, with $0$ equivalent to no length-control.
Unless otherwise specified, we use the specific thought prompt trained using the ArmoRM judge on UltraFeedback instructions as the default setup. 

As a baseline, we train the same seed model that outputs responses directly without any thinking (note, this can still perform CoT as a part of the response due to its initial instruction training).
We train this baseline in the exactly same way, using the same judge, data and loss.
This allows us to directly measure the effect of the thoughts on response quality.

For evaluation, we use two public benchmarks that test general instruction following capability: AlpacaEval 2 \citep{dubois2024length} and Arena-Hard \citep{li2024crowdsourced}.
Both benchmarks perform auto-evaluation using gpt-4-1106-preview as a judge.
In addition, we also create our own benchmark to perform fine-grained analysis.

\begin{table}[t]
    \centering
    \begin{tabular}{lcc}
        \toprule
         \bf Method & \bf AlpacaEval (LC) & \bf Arena-Hard \\
         \midrule
         {\small \hspace{-2mm} \emph{Llama-3-8B-Instruct-based} } \\
         Llama-3-8B-Instruct & 24.9 & 20.6 \\
         Llama-3-8B-Instruct + Thought prompt & 17.3 & 14.1 \\
         Direct response baseline & 48.4 & 33.0 \\
         \methodshort{} & \bf 52.5 & \bf 37.3 \\
         \midrule
         {\small \hspace{-2mm} \emph{Larger models} } \\
         GPT-4 (06/13) & 30.2 & 37.9 \\
         Llama-3-70b-instruct & 34.4 & 46.6 \\
         Mistral Large (24/02) & 32.7 & 37.7 \\
         Qwen2 72B Instruct & 38.1 & 36.1 \\
         \bottomrule
    \end{tabular}
    \caption{{\bf Benchmark win rates (\%) for AlpacaEval (length-controlled (LC)) and Arena-Hard.}  We compare our method Thought Preference Optimization (TPO) to the direct response baseline, Llama-3-8B-Instruct, and Llama-3-8B-Instruct using Thought Prompting. The latter, which does not perform well, is used as initialization for the first iteration of TPO training. TPO optimizes thought generation during iterative training, which then outperforms the baselines.  
    We also include several well-known LLMs as  reference which are typically larger than our TPO model. 
    }
    \label{tab:summary}
\end{table}

\subsection{AlpacaEval Results}

\begin{figure}[t]
    \centering
    \hfill
    \includegraphics[scale=0.7]{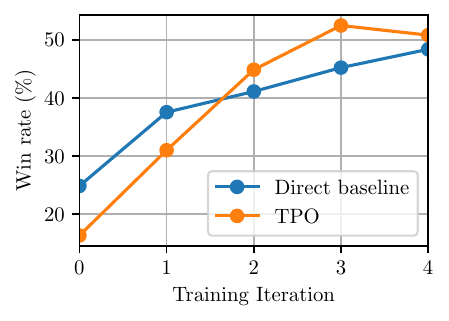}
    \hfill
    \includegraphics[scale=0.7]{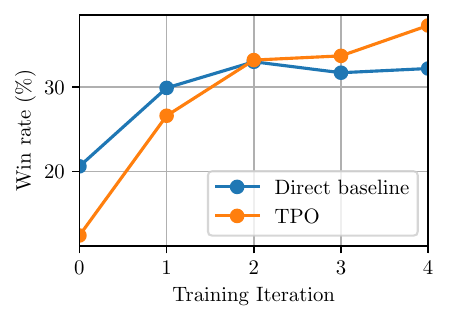}
    \hfill
    \caption{Training iterations on AlpacaEval (left) and Arena-Hard (right), comparing our TPO method to the direct baseline starting from the seed (iteration 0) model.
    }
    \label{fig:alpaca}
\end{figure}

The highest win rate our model \methodshort{} achieves is 52.5\%, which is +4.1\% better than the direct baseline, as shown in \autoref{tab:summary}.
It is also a +27.6\% increase over the seed model and puts our method in 3rd position on the leaderboard\footnote{As of Sep. 27th 2024. \url{https://tatsu-lab.github.io/alpaca_eval/}}, just after GPT-4 Omni and GPT-4 Turbo.
This is an impressive result given the small size (8B) of our model. 

In \autoref{fig:alpaca}, we plot the win rate for different iterations of training.
We can see that before training (iteration 0) the direct baseline performs much better.
This is expected as the seed model is instruction tuned to directly output a response.
Simply prompting the model to write down its thought processes actually hurts performance.
This agrees with the findings by \citet{sprague2024cot} who showed CoT prompting only helps math and logic related tasks.

However, after several iterations of training \methodshort{} catches up with the baseline, even outperforming it in iteration 3 and 4.
This is a promising indication that the model is adapting to think in a way that uses those thoughts to improve its responses.
A more detailed breakdown of results is given in \autoref{tab:alpaca}.
The best performing setup uses UltraFeedback instructions for training with an ArmoRM judge for both TPO and direct models.
For \methodshort{}, both generic and specific thought prompts produced similar outcomes, but the latter performs slightly better (both \autoref{tab:summary} and \autoref{fig:alpaca} use this setup).

\begin{table}[t]
    \centering
    \caption{Breakdown of AlpacaEval results for different judge models, training instruction sets (``Data'') and Thought Prompts comparing our method, TPO, to the direct baseline. We report the standard LC win rate.
    In each case TPO outperforms the direct baseline, but only after several iterations of training.
    }
    \label{tab:alpaca}
    \vspace{0.5em}
    \begin{tabular}{lcccccccc}
         \toprule
                \multicolumn{4}{c}{\bf Training Setup} & \multicolumn{5}{c}{\bf Training Iteration}\\
                \cmidrule(lr){1-4}\cmidrule(lr){5-9}
          \bf Model &  \bf Judge & \bf Data & \bf Thought Prompt & \bf  0 & \bf 1 & \bf 2 & \bf 3 & \bf 4 \\  
         \midrule
         Direct  &  STE & Synthetic &   -  & 24.9 & 34.8 & 40.3 & 43.6 & 42.6 \\
         TPO     &  STE & Synthetic & Generic & 17.3 & 32.0 & 40.2 & \textbf{47.6}& 46.8  \\
         \midrule
         Direct  &  Armo & Synthetic &   -     & 24.9 & 35.3 & 43.2 &  45.7 & 48.1 \\
         TPO     &  Armo & Synthetic & Generic & 17.3 & 33.8 & 44.1 & 49.8 & {\bf 51.1} \\
         \midrule
         Direct  &  Armo & {\small UltraFeedback} &   -  &   24.9   & 37.6 & 41.1 & 45.2       & 48.4 \\
         TPO     &  Armo & {\small UltraFeedback}  & Generic &  17.3  & 32.8 & 43.0 & 47.0       & 50.2 \\
         TPO     &  Armo & {\small UltraFeedback}  & Specific & 16.4 & 31.0 & 44.9 & {\bf 52.5} & 50.8 \\
         \bottomrule
    \end{tabular}
\end{table}

\subsection{Arena-Hard Results}

Results on the Arena-Hard benchmark are shown  in \autoref{tab:summary} (right) and \autoref{fig:alpaca} (right).
They follow a similar trend to the results from AlpacaEval.
Thinking performance is poor with the initial seed model at the start of training, but with more training iterations it matches the direct baseline and starts to outperform it.
\methodshort{} reaches a win rate of 37.3\%, which is +4.3\% better than the baseline.
This makes our model the best model on the leaderboard with a such small size\footnote{Excluding some closed models like GPT-4 where the size is not disclosed. Compared against Aug. 31st 2024 version. \url{https://github.com/lm-sys/arena-hard-auto}}. It performs similarly to much larger models like GPT-4 (06/13) or Mistral Large (24/02).
Detailed results comparing different experimental setups are shown in Appendix \autoref{tab:arena_detail}.

\subsection{Fine-grained Evaluation}
\begin{figure}[t]
    \centering
    \includegraphics[width=0.98\linewidth]{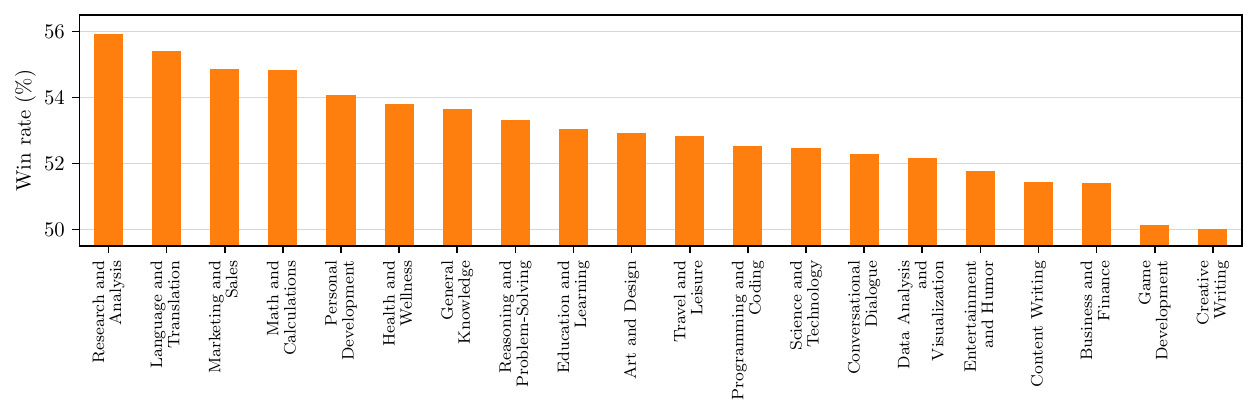}
    \caption{{\bf Fine-Grained evaluation on unseen instructions from UltraFeedback, broken down by category.}
    We measure the win rate of \methodshort{} against the direct baseline as judged by GPT4.
    }
    \label{fig:finegrain_wildchat}
\end{figure}

While the above benchmarks evaluate overall performance, they lack granularity to inform which types of instructions benefit from thinking.
To obtain a more fine-grained evaluation, we build our own evaluation using UltraFeedback.
We take instructions not used in training, and assign them individually to one of 20 categories until each category has 200 samples. 

To measure the performance on this dataset, we compare responses generated from  our \methodshort{} model against responses from the direct baseline model.
We use the evaluation prompt from Arena-Hard and GPT4 (gpt-4-1106) as a judge.
The ordering of responses is randomized to reduce position-bias. 
\autoref{fig:finegrain_wildchat} shows the win rates of \methodshort{} on all 20 categories.
Surprisingly, we observe that non-reasoning categories obtain  large gains through thinking.
This includes language and translation, marketing and health.
We also see improvement in reasoning categories like research and analysis, math and calculations.
See \autoref{app:finegrained_details} for more details and additional results.

\subsection{Analysis and Ablations}
In this section, we present analysis and several ablation results to further understand \methodshort{}.

\paragraph{Qualitative analysis of thoughts}
In \autoref{fig:poem}, we present an example of a non-reasoning task to illustrate the broader utility of thinking.
While writing a poem is not typically viewed as a reasoning task, it can benefit from better planning and understanding of the instruction.
This is also a good example of when it makes sense to hide the thought process.
\autoref{fig:specific_example} shows the thinking of a TPO model trained with the specific thought prompt on a factoid question (which dog breed is smallest).
We see that the model thinks first about the question, then evaluates its draft response.
We provide more thought examples in \autoref{sec:examples}.

\begin{figure}[t]
    \centering
    \begin{prompt}{Example of Non-Reasoning Instruction (generic thought prompt)}
Instruction: Write me a poem in the style of neruda

\tcblower

{\color{ForestGreen}
\textbf{Here is my thought process:}
\begin{itemize}[leftmargin=0.3cm,itemsep=0mm]
\item Neruda's style is known for being passionate, sensual, and evocative. I need to tap into that emotional depth.
\item His poetry often explores themes of love, politics, and social justice. I should choose a theme that resonates with Neruda's concerns.
\item Neruda's writing is also infused with a sense of mysticism and the mysterious. I'll try to incorporate some enigmatic elements.
\end{itemize}
\textbf{Here is my response:}
}\\
\\
\textbf{Whispers in the Rainy Night}
\\
In the city's secret heart, where shadows play

Like lovers' fingers tracing the city's gray

The rain whispers secrets, a nocturnal refrain

As I stand here, lost in your eyes' sweet pain \emph{(continued ...)}

\end{prompt}
    \caption{Example of \methodshort{} responding to a non-reasoning instruction to write a poem. This example demonstrates how thinking can be useful in understanding the task and planning accordingly even for creative writing. The thought part in green is not part of the response.}
    \label{fig:poem}
\end{figure}
\begin{figure}[t]
    \centering
    \begin{prompt}{Example of Factoid Question (specific thought prompt)}
Instruction: What breed dog is smallest?

\tcblower
{\color{ForestGreen}

\textbf{Internal Thoughts}
\vspace{1.5mm}

``Okay, let's tackle this query. I'm thinking of popular toy breeds that are widely recognized for their small size. Chihuahua, Pomeranian, and Poodle are some of the breeds that come to mind. But, I need to dig deeper to ensure I provide the most accurate answer.

\vspace{1.5mm}
Draft Response:
\vspace{1.5mm}

The smallest dog breed recognized by kennel clubs and breed associations is the Chihuahua, specifically the Long Coat variety. According to the FCI (Fédération Cynologique Internationale), the Chihuahua has an average weight range of 1.5-3 kg (3.3-6.6 lbs) and an average height range of 15-23 cm (6-9 inches).

\vspace{1.5mm}
Evaluation:
\vspace{1.5mm}

My draft response looks good, but I'm not entirely satisfied. While Chihuahua is indeed a contender for the smallest breed, I'm not certain if it's the absolute smallest. I should provide more context and potentially mention other breeds that might be considered smaller. Additionally, I could include some interesting facts or tidbits to make the response more engaging.

\vspace{1.5mm}
\textbf{Final Response}
\vspace{1.5mm}

\textless R\textgreater
\vspace{1.5mm}
}

\textbf{The Smallest Dog Breed: Meet the Chihuahua}
\vspace{1.5mm}

While there might be variations in size within breeds, the Chihuahua is widely recognized as the smallest dog breed by kennel clubs and breed associations. According to the Fédération Cynologique Internationale (FCI), the Teacup Chihuahua (Long Coat variety) boasts an average weight of 1.5-3 kg (3.3-6.6 lbs) and an average height of 15-23 cm (6-9 inches). Some exceptional specimens can even reach a mere 10 inches (25 cm) in height! Despite their tiny size, Chihuahuas pack a big personality punch, making them a beloved companion for many dog enthusiasts worldwide.

\end{prompt}
    \caption{Example of \methodshort{} answering a simple factoid question.
    This model is trained with the specific thought prompt, so it writes a draft and evaluates it in the thought part (in green).
    }
    \label{fig:specific_example}
\end{figure}

\paragraph{Thought Prompt Types}
\autoref{tab:inst_types} compares different thought prompt types in terms of different metrics across iterations.
As we have observed previously, the seed model performs poorly when it is asked to generate thoughts compared to directly generating a response, which is the result of the model being well optimized to do the latter.
However, after one iteration of training, we see the gap is shrinking.
Between the two thought prompt types we try, there is not much difference in terms of win rate, but the thought and answer lengths vary greatly.
Also, the specific thought prompt leads to longer responses and have a higher average ArmoRM score (see Appendix
\autoref{fig:armo_reward}).

\paragraph{Thought lengths}
As shown in \autoref{tab:inst_types}, the specific thought prompt has the longest thought generations because it asks the model to generate a draft response and evaluate it within the thought.
Although we did not directly supervise the thought process, we found the model learns to shorten and condense the thought throughout the training.  
\autoref{tab:inst_types} shows the length changes after 1 training iteration, but more detailed statistics across training are given in Appendix \autoref{tab:thought_len}.
We find that after 4 iterations of training, the thought part is shrunk by 61\% and 30\% for the generic and specific thought prompts respectively.
We also measure thought lengths by category (Appendix \autoref{fig:fine_len}) and find that categories such as research, art and writing have longer thoughts, while conversation, language, math and general knowledge are on the opposite end. 
\begin{table}[t]
    \centering
    \caption{Comparing the generic and specific (\autoref{prompt:std}) thought prompts.
    We measure the length (in characters) of the thought and final answer on the UltraFeedback train prompts, as well as AlpacaEval (LC) win rates before and after one training iteration.}
    \label{tab:inst_types}
    \vspace{0.5em}
    \begin{tabular}{lrrrrrr}
         \toprule
         & \multicolumn{3}{c}{\bf Length of the seed model} & \multicolumn{3}{c}{\bf Length after iter 1} \\
         \cmidrule(lr){2-4}\cmidrule(lr){5-7}
         \bf Thought Prompt & \bf Thought & \bf Response & \bf Win rate 
         & \bf Thought & \bf Response & \bf Win rate \\
         \midrule
         None (Direct)  &  - & 2878 & 24.9\% & - & 2787 & 37.5\% \\ 
         Standard & 799 & 1717 & 17.3\% & 606 & 1638 & 32.7\%\\
         Specific & 2193 & 1300 & 16.4\% & 1613 & 1431 & 31.0\%\\
         \bottomrule
    \end{tabular}
\end{table}

\paragraph{Length-Control}
We find that length-control (LC) is a must during training, especially with the \eft{} judge.
When no LC is performed, the average response length grows rapidly \citep{Wu2024MetaRewardingLM}.
For instance, after just 1 iteration of training of the direct baseline, the average response length grows around 15\%.
If we utilize our LC technique when building preference pairs, we can maintain the same length or even decrease the length if necassary.
In our experiments, we tune the LC coefficient $\rho$ by measuring  the average length of chosen and rejected samples, and choose the smallest value such that the former is not longer than the latter.
This simple method was very effective, and both \methodshort{} and the direct baseline model did not grow their response length much during training, as demonstrated in \autoref{tab:inst_types}.

\paragraph{Training Instruction set and Judges}
We experiment with two training instruction sets: the synthetic instruction set generated by few-shot prompting Llama-2-70B, and UltraFeedback - a more curated instruction set that contains human instructions. 
As shown in \autoref{tab:alpaca} and  Appendix \autoref{tab:arena_detail}, we do not observe a large difference between these two datasets in terms of final performance on the benchmarks.
We also experimented with two different judges for training, ArmoRM and the \eft{} judge. 
We find that when using either judge  \methodshort{} outperforms the direct baseline on AlpacaEval as shown in \autoref{tab:alpaca}.
However, \eft{} required more length-control to be applied during training.
On both benchmarks, we obtained better results with the ArmoRM judge.

\paragraph{Parsing errors}
For the initial seed model we use prompting to maintain the desired output format, where thoughts and answer are separated, see \autoref{fig:prompt}. However, we find during training that without additional safeguards, the rate of parse errors keep fluctuating, and is sometimes hard to control. In order to teach the model to follow the constraint, we add responses with parse errors as rejected examples in the preference data creation process. Typically, the ratio of parse errors as rejected is no more than 10\% in each iteration, but we observe for the specific prompt template, it is much harder to control. 
We found that sampling again if a parse error occurred helped alleviate the issue during evaluations.

\paragraph{DPO vs IRPO loss}
In addition to training with the DPO loss, we also experimented the IRPO loss that has shown promising results in math tasks \citep{pang2024iterative}.
IRPO adds an NLL loss on the chosen samples, so that their average log probability does not decrease, which can occur with DPO.
However, in our experiments and setting, we did  not see a noticeable difference in performance.
After one training iteration, \methodshort{} using IRPO  gives a 31.6\% AlpacaEval LC win rate, which almost matches the 32.0\% of DPO.
Here we used the \eft{} judge and the synthetic dataset for training.

\paragraph{Math Domain}
To further understand the performance on the math domain where CoT techniques are often applied, we evaluate our model on the GSM8K dataset \citep{cobbe2021gsm} that contains grade-school math word problems.
Since we have the correct answers in this dataset, we can more accurately measure performance compared to relying on a judge.
We first test the seed model without a thought prompt, but observe it uses CoT anyway due to its instruct training, obtaining 79.2\% accuracy. We thus append ``Output only the number answer.'' to the problems so that the model will not perform CoT in its response.
It still performs surprisingly well in this answer-only mode, reaching 69.7\% accuracy.
However, on closer inspection we noticed that the seed model still performs CoT-like arithmetic operations like ``16 - 3 - 4 = 9. 9 * 2 = 18'' in its output.
We then evaluate direct models trained on the synthetic prompts and the \eft{}  judge. After training, this number drops to 51.3\%. Performance is even lower for TPO models.
This indicates that our experimental setup is not suited for math tasks.
In fact, only 2.2\% of our training instructions are categorized into the math category (see Appendix \autoref{fig:ultra_count}).
Such a drop in math performance is also observed by \citet{meng2024simpo,zhu2024starling} in a general instruction tuning setup.
However, we do see some examples where the TPO model is able to correct its mistake through reflection, while the direct model simply stops at the mistake, as shown in Appendix \autoref{fig:gsm8k_tpo_self_correct}.
For more detailed results, see Appendix \autoref{tab:gsm8k}.


\section{Related work}
\label{sec:related}

\paragraph{Reasoning in Language vs. Vectors:}
In this work we focus on thinking that is explicitly in natural language. 
Thinking in words takes advantage of the natural language understanding capability of LLMs. LLMs are trained on large pretraining corpora of human text, which contain human thoughts expressed in natural language, and this thinking ability is hence encoded into the model.
While thinking in continuous values might provide more bandwith, the Transformer architecture already can compute continuous vectors as hidden states and feed them to the next layer.
However, these hidden vectors do not feed back into the model at the next token, and thus are not accessible to the future lower layers \citep{fan2020addressing}.
Word tokens on the other hand are fed back to the model immediately, i.e. during inference the previous output token is fed as input to predict the next token --
making it possible to condition all future computations on them \citep{merrill2023expresssive}.
Another advantage of word tokens is that there exist simple sampling mechanisms which allow thoughts to take different paths each time \citep{wang2022self}, which can be used to improve results e.g. via majority vote.

\paragraph{Chain-of-Thought (CoT):}
CoT prompting \citep{wei2022chain} demonstrated that LLMs perform better at reasoning tasks when they are encouraged to write down intermediate reasoning steps.
Since the type of thinking in CoT is dictated by the prompt instruction, there are now many different variations of it facilitating different types of reasoning, such as decomposing into smaller problems \citep{DBLP:conf/iclr/ZhouSHWS0SCBLC23}.
It is now widely used for math and reasoning tasks, and most current LLMs are finetuned to do CoT by default for those types of tasks \citep{dubey2024llama}.
Other works like \citet{pfau2024let} show that the model equipped with CoT might be able to perform hidden thinking using filler tokens.
However, CoT usage has had more limited use in other types of tasks.
Meta-analysis by \cite{sprague2024cot} found that CoT techniques have little benefit outside of math and logic related tasks. 

\paragraph{Training to Think:}
There have been other previous efforts to train LLMs to think. 
\citet{Nye2021ShowYW} trained a model to write intermediate calculations into a scratchpad section before writing the final answer, which improved performance in math and coding tasks.
Similarly \citet{Lehnert2024BeyondAB} showed that Transformers can solve complex planning tasks if they are trained to write A* search traces before outputting the solution.
However, these methods rely on supervised training so ground-truth thought data is required.
STaR \citep{zelikman2022star} removes this constraint by generating both thought and answer from a model using few-shot prompting.
Then the generations are filtered by the correctness of the answer to be used for supervised finetuning.
It also has an option to feed correct answers to the model to generate better thought candidates.
It was applied to multi-choice reasoning and math tasks where the correct answers were available.
Its generalization Quiet-STaR \citep{zelikman2024quietstar} aims to insert thought segments into unstructured text.
This involves sampling a sequence of thought tokens after every input token, then training using a REINFORCE based loss that optimizes the likelihood of subsequent input tokens.
While it showed promising results in multi-choice reasoning and math tasks, the training mechanism is complex and compute heavy. 
V-STaR \citep{hosseini2024v} trained a DPO verifier on both correct and incorrect solutions and uses the verifier to select the response in inference time.
IRPO \citep{pang2024iterative} also trains a variant of DPO on math and reasoning problems to learn CoTs, assuming access to gold labels on the training set.
Similarly, Self-Notes \citep{lanchantin2023learning} allows the model to deviate from the input at any time to write its thoughts, 
but relied on supervised training data in symbolic tasks.
None of these methods have been applied to general instruction following using LLMs.

\paragraph{System 2 methods:} Many system 2 methods emerged in recent years that add intermediate steps at inference time before producing a final answer.
Those steps typically involve prompting the model with a certain goal, such as verification of the answer \citep{DBLP:conf/acl/DhuliawalaKXRLC24}, rephrasing user questions \citep{deng2023rephrase}, selecting sentences to attend to \citep{weston2023system}, etc.
\citet{briakou2024translating} developed a method for translation incorporating intermediate steps of drafting and revising.
Our \methodshort{} method has a similarity with these methods in the first step because it uses prompting on the initial seed model, but then optimizes the thoughts during training iterations. 
In contrast, the common feature of the system 2 methods just described is their reliance on hand-crafted prompts designed for a specific goal (e.g. verification), without optimizing those steps via finetuning.
Concurrent work by \citet{Kumar2024TrainingLM} trains models to self-correct, while \citet{yu2024distilling} distill system 2 methods 
into system 1 with supervised finetuning.
Rather than focusing on general thinking, these works teach the model specific skills.

\section{Conclusion}
\label{sec:conclustion}
In this paper, we introduced 
Thinking LLMs, which think in natural language before writing a response for {\em general instruction-following tasks}.
To train such models,  we proposed a new training recipe called \method{} for teaching Thinking LLMs to improve their thoughts.
Unlike prior methods \citep{snell2024scaling,kumar2024training}, which directly supervise the thought generation process through techniques like self-correction or self-refinement, we instead {\em provide incentives for the model to generate its own thoughts}, without explicitly teaching it how to think.
In our experiments, we train and evaluate the models in the general instruction following setup.
The results on benchmarks show that the initial seed model and first iterations of training of the Thinking LLM perform poorly compared to the typical direct response model.
However, after multiple iterations of training using  \methodshort{}, our method outperforms the  baseline.
Further, fine-grained evaluations reveal that thinking helps in categories that are not usually associated with reasoning or chain-of-thought methods.
This is an encouraging result and hopefully leads to wider adoption of Thinking LLMs in non-reasoning domains.

\section{Limitations}
\label{sec:limitations}
We experimented with two different thought prompts, and observed some performance differences between them.
It is likely that certain thought types are suited for certain tasks, and direct responses would even work better in certain situations.
Therefore, training on a diverse set of thought prompts and allowing the model to switch between them could potentially lead to further improvements in  performance. This would allow the model to better search the space of possible thoughts in order to learn to choose the most appropriate ones. However, we have not conducted these experiments.

While we see improvement in overall performance with \methodshort{}, 
evaluation on GSM8K showed degraded math performance.
As we discussed, this is likely due to our setup not being oriented toward such tasks.
Incorporating more math instructions during training and having access to a judge capable of evaluating of their answers are likely solutions.

In the current version of the method, thought lengths are purely determined by model itself.
There is no steerability in terms of changing the number of thought tokens.
Adding such functionality could be useful as longer thoughts increase computation and corresponding cost per user instruction.
We could use techniques like \citet{yuan2024following} for this purpose. 

All our experiments are based on 8B parameter sized models.
However, it is worth investigating the effect of thinking on larger scale models.
Given the compute requirements of such experiments, we leave that to future work.


\bibliography{iclr2024_conference}
\bibliographystyle{iclr2024_conference}

\clearpage
\appendix



\begin{table}[t]
    \centering
    \caption{Breakdown of Arena-Hard results for varying experimental setups. We report the default win rates.
    }
    \label{tab:arena_detail}
    \vspace{0.5em}
    \begin{tabular}{lcccccccc}
         \toprule
                &     &      &                & \multicolumn{5}{c}{\bf Training Iteration}\\
                \cmidrule{5-9}
          \bf Model &  \bf Judge & \bf Data & \bf Thought Prompt &  \bf  & \bf 1 & \bf 2 & \bf 3 & \bf 4 \\  
         \midrule
         Direct  &  STE & Synthetic &   -  & 20.6 & 29.6 & 32.5 & 33.1 & \bf 33.2 \\
         TPO     &  STE & Synthetic & Generic & 14.1 & 24.0 & 30.6 & 27.2 & 24.3 \\
         \midrule
         Direct  &  Armo & Synthetic &   -     &  20.6  & 27.1 & 31.4 & 31.1 & 30.4 \\
         TPO     &  Armo & Synthetic & Generic & 14.1& 25.7 & 31.0 & \bf 32.8 & 30.1 \\
         \midrule
         Direct  &  Armo & {\small UltraFeedback} &   -  & 20.6  & 29.9 & 33.0 & 31.7 & 32.2 \\
         TPO     &  Armo & {\small UltraFeedback}  & Generic & 14.1 & 27.2 & 28.1 & 30.8 & 31.8 \\
         TPO     &  Armo & {\small UltraFeedback}  & Specific & 12.4 & 26.6 & 33.2 & 33.7 & \bf 37.3 \\
         \bottomrule
    \end{tabular}
\end{table}

\section{Fine-grained evaluation details}
\label{app:finegrained_details}
\begin{figure}[t]
    \centering
    \includegraphics[width=1\linewidth]{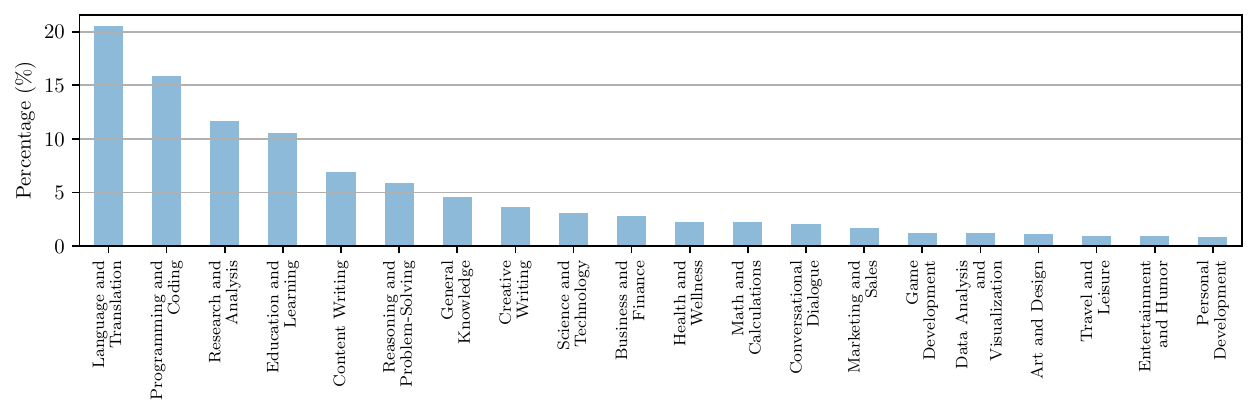}
    \caption{Training instruction distribution: we categorize  our training instructions into 20 topics. The language and translation category has the most examples, while the math category is only 2.2\% of the data.}
    \label{fig:ultra_count}
\end{figure}

\begin{figure}[t]
    \centering
    \small
    \includegraphics[width=0.97\linewidth]{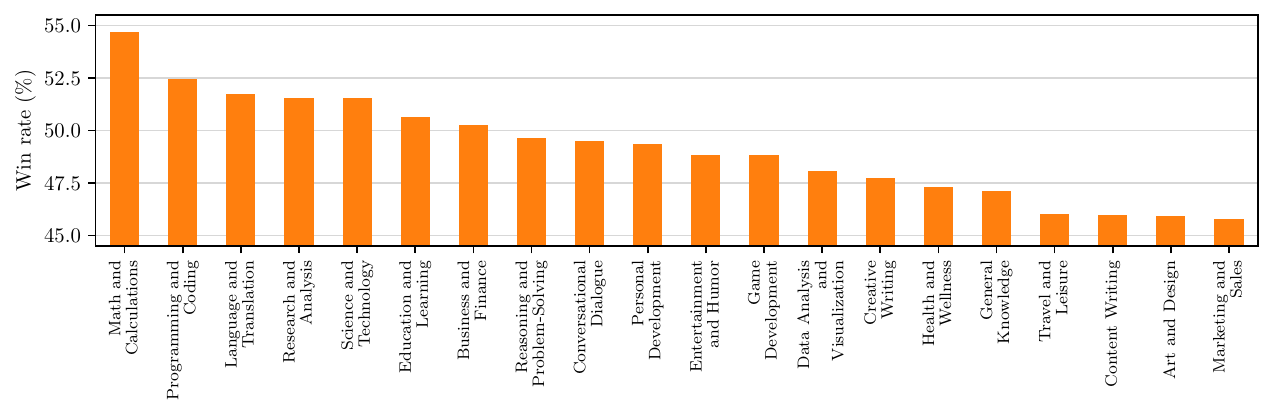}\\
    (a) Generic Thought Prompt judged by GPT4\\
    \includegraphics[width=0.97\linewidth]{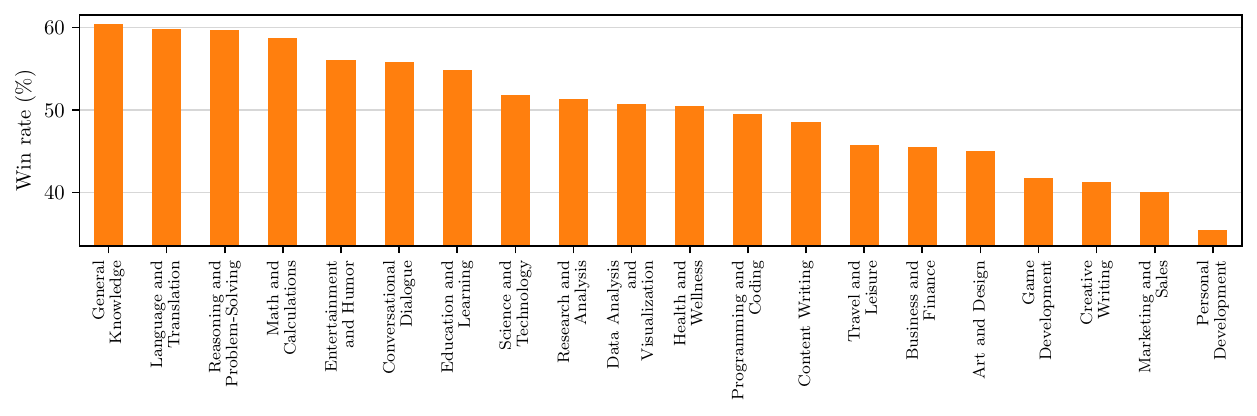}\\
    (b) Specific Thought Prompt judged by \eft{}
    \includegraphics[width=0.97\linewidth]{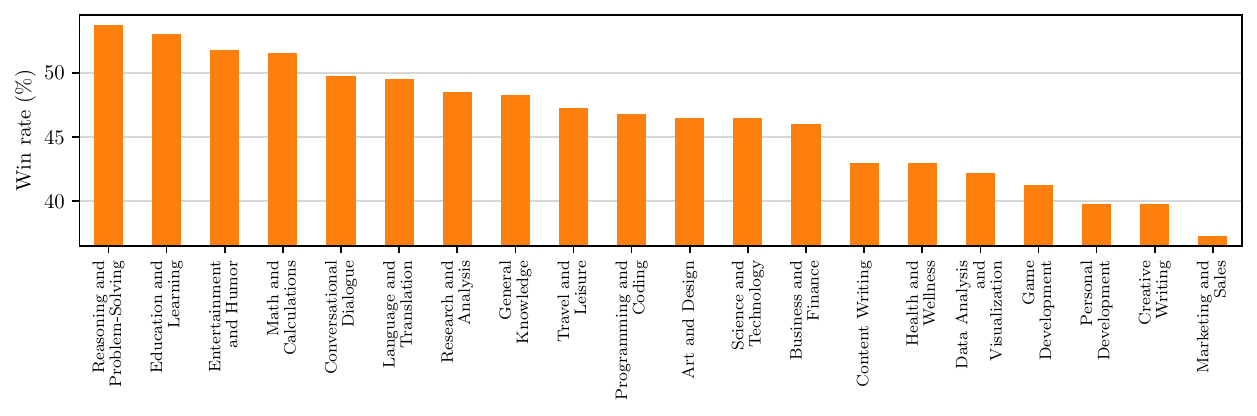}\\
    (c) Generic Thought Prompt judged by \eft{}\\
    \includegraphics[width=0.97\linewidth]{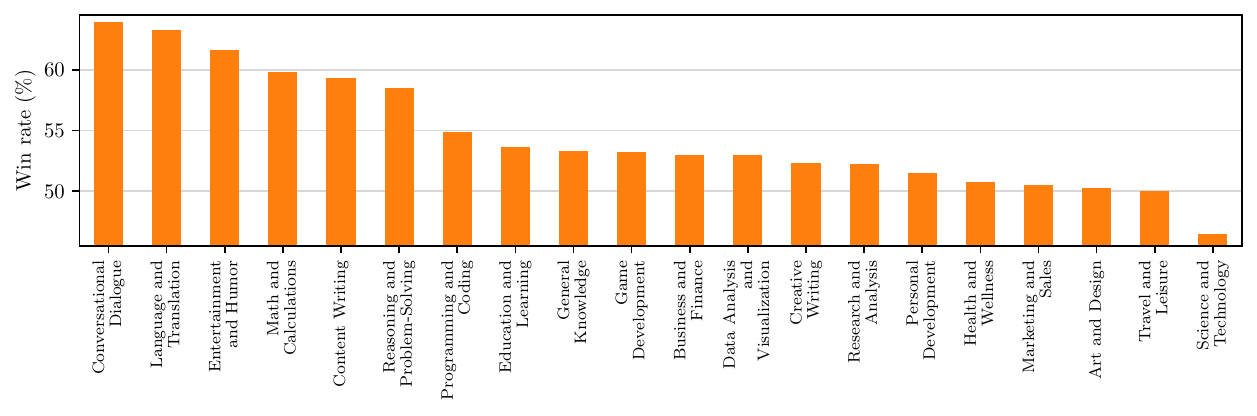}\\
    (d) Generic Thought Prompt judged by \eft{} when \eft{} is also used during training (iteration 3)\\

    \caption{\textbf{Additional Fine-grained evaluation results.} We show the win rate of \methodshort{} against the direct baseline trained in the same setup. The different plots correspond to setups differing in evaluation judge model, thought prompt type, and the training judge.}
    \label{fig:fine_grain_more}
\end{figure}

\begin{figure}[t]
    \centering
    \includegraphics[width=0.98\linewidth]{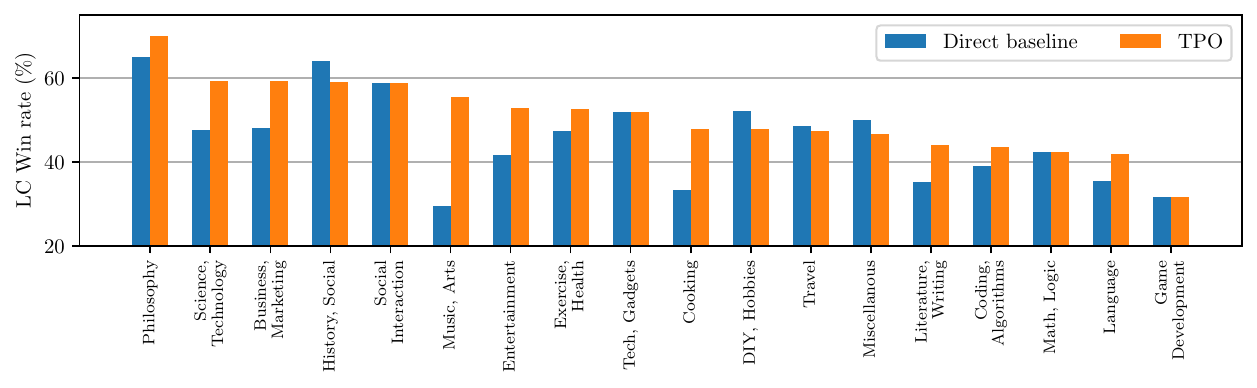}
    \caption{Fine-grained evaluation on AlpacaEval instructions. We measure length-controlled win rates within each category for each model.}
    \label{fig:alpaca_cat2}
\end{figure}

We first generate 20 category names by prompting Llama-3.1-70B-Instruct model to summarize the categories present given many instructions from WildChat \citep{zhao2024wildchat}. 
We select 200 instructions from UltraFeedback per category for evaluation, excluding those that were used in training. 
\autoref{fig:ultra_count} shows the ratio of each category in our training instructions from UltraFeedback.
This shows that some categories have much fewer samples in the training data, which might negatively affect their performance.

We show the prompts we used: to generate categories in \autoref{fig:prompt_gencat}, assign categories in \autoref{fig:prompt_assign}, and evaluate with the GPT4-1106 judge in \autoref{fig:prompt_eval_gpt4}.
The judge can give evaluations of: A\textgreater\textgreater  B, A\textgreater B, A=B, B\textgreater A, B\textgreater\textgreater  A, which are counted as  $1, 0.75, 0.5, 0.25, 0$ respectively. 
The scores are averaged over instructions to determine the win rate in the category.
\autoref{fig:fine_grain_more} (a) shows the win rate of the generic thought prompt, which is lower in average compared to the specific thought prompt (see \autoref{fig:finegrain_wildchat}).
However, it still outperforms the direct baseline on complex categories like math and coding.

\begin{figure}
    \centering

\begin{prompt}{Prompt to come up with categories}
Given the following list of possible instructions, define a maximum of 20 categories that would cover the types of intructions, for example recipes, reasoning tasks, general knowledge etc. Try to cover as many of the instructions as possible with the maximum 20 categories, while keeping the categories high-level, concise, simple, and easy to understand.\\

\textless NEW INSTRUCTION\textgreater\\
{\color{NavyBlue}\{query\}}\\

\textless NEW INSTRUCTION\textgreater\\
{\color{NavyBlue}\{query\}}\\

...

\end{prompt}
    \caption{Prompt to generate 20 categories based on a set of user instructions.}
    \label{fig:prompt_gencat}
\end{figure}

\begin{figure}
    \centering
\begin{prompt}{Prompt to assign categories}
Below is an instruction that I would like you to analyze:\\

\textless instruction\textgreater \\
{\color{NavyBlue}
\{user instruction\}}\\
\textless /instruction\textgreater \\

Categorize the instruction above into one of the following categories:\\
General Knowledge\\
Math and Calculations\\
Programming and Coding\\
Reasoning and Problem-Solving\\
Creative Writing\\
Content Writing\\
Art and Design\\
Language and Translation\\
Research and Analysis\\
Conversational Dialogue\\
Data Analysis and Visualization\\
Business and Finance\\
Education and Learning\\
Science and Technology\\
Health and Wellness\\
Personal Development\\
Entertainment and Humor\\
Travel and Leisure\\
Marketing and Sales\\
Game Development\\
Miscellaneous\\

Be sure to provide the exact category name without any additional text.

\end{prompt}
    \caption{Prompt to assign one of the 20 categories to user instructions}
    \label{fig:prompt_assign}
\end{figure}






\begin{figure}
    \centering
\begin{prompt}{Prompt to evaluate responses with GPT4}
{\color{Gray}\textit{\textless system\textgreater}}\\

Please act as an impartial judge and evaluate the quality of the responses provided by two AI assistants to the user prompt displayed below. You will be given assistant A's answer and assistant B's answer. Your job is to evaluate which assistant's answer is better.\\

Begin your evaluation by generating your own answer to the prompt. You must provide your answers before judging any answers.\\

When evaluating the assistants' answers, compare both assistants' answers with your answer. You must identify and correct any mistakes or inaccurate information.\\

Then consider if the assistant's answers are helpful, relevant, and concise. Helpful means the answer correctly responds to the prompt or follows the instructions. Note when user prompt has any ambiguity or more than one interpretation, it is more helpful and appropriate to ask for clarifications or more information from the user than providing an answer based on assumptions. Relevant means all parts of the response closely connect or are appropriate to what is being asked. Concise means the response is clear and not verbose or excessive.\\

Then consider the creativity and novelty of the assistant's answers when needed. Finally, identify any missing important information in the assistants' answers that would be beneficial to include when responding to the user prompt.\\

After providing your explanation, you must output only one of the following choices as your final verdict with a label:\\

1. Assistant A is significantly better: [[A\textgreater \textgreater B]]\\
2. Assistant A is slightly better: [[A\textgreater B]]\\
3. Tie, relatively the same: [[A=B]]\\
4. Assistant B is slightly better: [[B\textgreater A]]\\
5. Assistant B is significantly better: [[B\textgreater \textgreater A]]\\

Example output: "My final verdict is tie: [[A=B]]".\\

{\color{Gray}\textit{\textless /system\textgreater}}\\


\textless $\vert$User Prompt$\vert$\textgreater \\
{\color{NavyBlue}\{prompt\}}\\

\textless $\vert$The Start of Assistant A's Answer$\vert$\textgreater\\
{\color{NavyBlue}\{generation\}}\\
\textless $\vert$The End of Assistant A's Answer$\vert$\textgreater\\

\textless $\vert$The Start of Assistant B's Answer$\vert$\textgreater\\
{\color{NavyBlue}\{generation2\}}\\
\textless $\vert$The End of Assistant B's Answer$\vert$\textgreater

\end{prompt}
    \caption{Prompt to evaluate using GPT4 judge in the fine-grained evaluation.}
    \label{fig:prompt_eval_gpt4}
\end{figure}

\begin{figure}
    \centering
\begin{prompt}{Prompt to evaluate responses with \eft{} judge}
Please act as an impartial judge and evaluate the quality of the responses provided by two AI assistants to the user question displayed below. You should choose the assistant that follows the user's instructions and answers the user's question better. Your evaluation should consider factors such as the helpfulness, relevance, accuracy, depth, creativity, and level of detail of their responses. Begin your evaluation by comparing the two responses and provide a short explanation. Avoid any position biases and ensure that the order in which the responses were presented does not influence your decision. Do not allow the length of the responses to influence your evaluation. Do not favor certain names of the assistants. Be as objective as possible. After providing your explanation, output your final verdict by strictly following this format: ``[[A]]'' if assistant A is better, ``[[B]]'' if assistant B is better.\\

\lbrack User Question\rbrack \\
{\color{NavyBlue}
\{input\} }\\

\lbrack The Start of Assistant A's Answer\rbrack \\
{\color{NavyBlue}\{generation\}}\\
\lbrack The End of Assistant A's Answer\rbrack \\

\lbrack The Start of Assistant B's Answer\rbrack\\
{\color{NavyBlue}\{generation2\}}\\
\lbrack The End of Assistant B's Answer\rbrack

\end{prompt}
    \caption{Prompt to evaluate using \eft{} judge in the fine-grained evaluation.}
    \label{fig:prompt_eval_eft}
\end{figure}



To further validate our results, we perform a similar evaluation using an alternative judge, \eft{}.
The prompt we used with this judge is shown in \autoref{fig:prompt_eval_eft}.
Here, we test 500 instructions per category and evaluate in both orders. 
\autoref{fig:fine_grain_more} (b) and (c) show the win rate of the specific and generic thought prompts respectively.
Again, we see that the specific thought prompt works better.
Compared to the GPT4-based evaluations, we see more variation across the categories, perhaps due to the lack of a tie option in the judge.
We also see that \methodshort{} does better when the training also uses the \eft{} judge (\autoref{fig:fine_grain_more} (d)).

Overall, we observe that ranking of categories by the win rate varies a lot between different judges and thought prompt types, making it challenging to draw a conclusion about which category benefits most from thinking.
However, several categories consistently improve from \methodshort{} such as math, language and translation, and reasoning.

In addition, we also classify AlpacaEval instructions into 20 categories following \cite{yuan2024selfrewarding} and present the default benchmark win rates in \autoref{fig:alpaca_cat2}.
We see large improvements in some categories like music and art, entertainment and cooking.
This is interesting because those categories are not considered as reasoning tasks in the standard literature.
However, note that some categories might not have sufficient samples (the total number of samples is only 805) even though we removed two categories that have less than 10 samples.

\begin{figure}
    \centering
    \includegraphics[width=\linewidth]{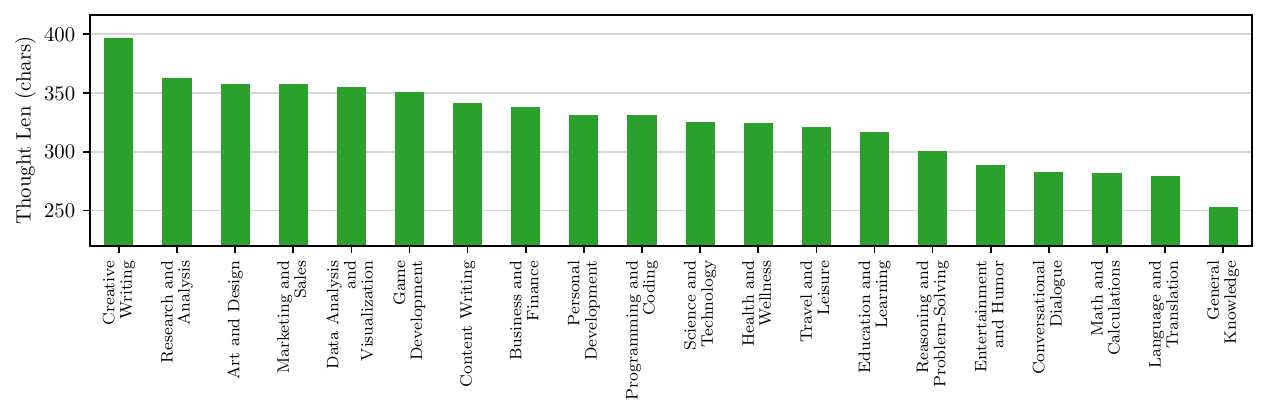}\\
    (a) Generic Thought Prompt
    \includegraphics[width=\linewidth]{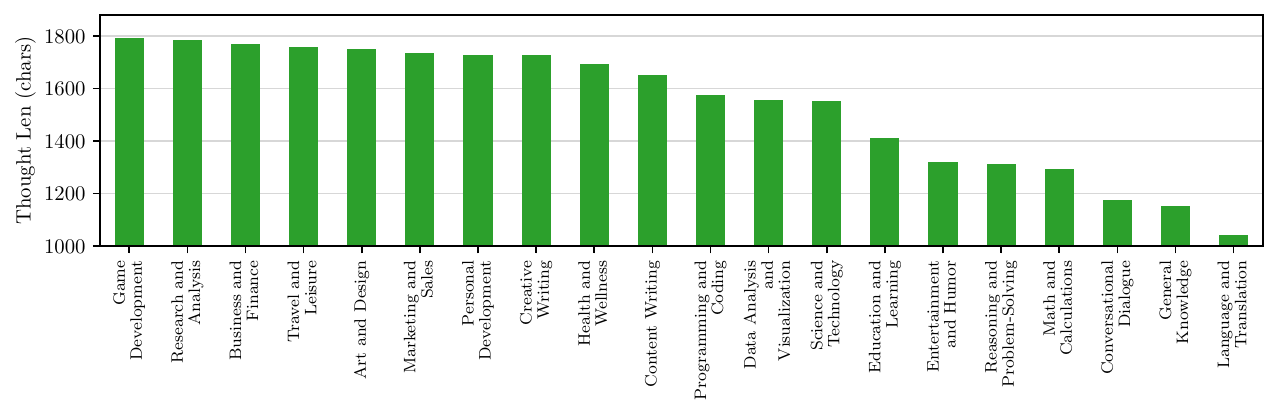}\\
    (b) Specific Thought Prompt
    \caption{Length of thoughts generated by \methodshort{} models.}
    \label{fig:fine_len}
\end{figure}

\section{ELO computation}
\label{elo_computation}

For each prompt $x$, we have up to $K$ corresponding responses denoted as $\{ y^1,\ldots,y^{K} \}$. We then evaluate each pair of different responses $(y^m, y^n)$ using the pairwise judge.

The result of a single battle between judgments $(y^m, y^n)$ is defined as:
\begin{align*}
r^{mn} = 
\begin{cases}
    1 & \text{If the pairwise-judge prefers } m\ \text{wins} \\
    -1 & \text{If the pairwise-judge prefers } n\ \text{wins} \\
    0 & \text{If tie or error} .
\end{cases} 
\end{align*}
We then construct a battle matrix $B$ as the combination of the battle results:
$$B_{mn} = \vone[r^{mn} = 1] + \vone[r^{nm} = -1]$$
The next step is to convert the battle matrix into ELO score. Following the same process as in \citet{Wu2024MetaRewardingLM}, we determine the ELO score $s^k$ for each response $y^k$ by solving the following maximum likelihood estimation problem:
$$\arg\max_{s^k}\sum_{1 \leq m,n \leq K} B_{mn}\log\left(\frac{e^{s^m-s^n}}{1+e^{s^m-s^n}}\right).$$

\section{Thought Examples}
\label{sec:examples}
Here we show several examples of thought and response outputs.
\autoref{fig:non_reasoning} is a non-reasoning question and \methodshort{} still takes advantage of thinking by recalling relevant information.
\autoref{fig:v11_win} is an example where the model solves complex problem by thinking of a draft that outlines the main steps before actually writing the code.
In contrast, the direct response baseline makes a mistake, as shown in \autoref{fig:baseline_fail}.

We do observe some failure cases of thinking.
For example, we observe the model learns to generate multiple round of the drafting-evaluation process even if it is not told to do so in the prompt. While this can be helpful, we find overthinking sometimes causes the model to get ``lost in the thought'', without giving an final answer or else providing a wrong answer. An example of this is shown in \autoref{fig:overthink}.

The next example shows how the model thinks before training  in \autoref{fig:good_cot_bad_addition}.
Here, the thought process is correct and helps to get closer to the answer, and the model only gets an addition wrong at the last step. 
This shows that thinking can be helpful, but the model is still bad at arithmetic.

\begin{figure}[t]
    \centering
    \begin{prompt}{Example of Non-Reasoning Instruction (generic thought prompt)}
Instruction: What type of soil is suitable for cactus?

\tcblower

{\color{ForestGreen}
\textbf{Here is my thought process:}
\begin{itemize}
\item  I need to think about the specific requirements of cacti to determine the suitable soil type.
\item  I recall that cacti are native to dry and arid regions, so the soil must be adapted to conserve water.
\item  I need to consider the pH level of the soil, as well.
\end{itemize}
\textbf{Here is my response:}
}
\\
Cacti thrive in well-draining soils that mimic their natural desert habitats. A suitable soil type for cactus cultivation typically possesses the following characteristics:
\begin{itemize}
\item[1] \textit{High drainage}: Cacti detest waterlogged soil, so a mix with excellent drainage is essential. A combination of perlite, vermiculite, or sand helps prevent root rot.
\item[2] \textit{Neutral to slightly alkaline pH}: Most cacti prefer a slightly acidic to neutral soil pH (6.0-7.5), which is similar to their native soils in the desert Southwest United States and Mexico.
\item[3] \textit{Low organic matter}: Cacti don't require a lot of nutrients, and excessive organic matter can retain too much water. A low-to-medium organic matter content (around 5-10\%) is suitable.
\item[4] \textit{Good aeration}: A well-balanced mix allows for air pockets and oxygen circulation around the roots, promoting healthy growth.
\end{itemize}
A general-purpose cactus potting mix typically contains a blend of:

\begin{itemize}
\item Coarse grit or perlite (40-60\%)
\item Peat moss or coconut husk (10-20\%)
\item Vermiculite (10-20\%)
\item Composted cactus mix or worm casting (5-10\%)
\item Small amounts of dolomitic limestone or crushed eggshells (pH adjustment and calcium supplementation)
\end{itemize}
You can also create your own cactus soil mix by combining these components in the above proportions. Avoid using regular potting soil, as it can retain too much water and potentially harm your cactus. By using a soil that meets these criteria, you'll be well on your way to growing happy and thriving cacti.

\end{prompt}
    \caption{Example of \methodshort{} answering a non-reasoning question. This example demonstrate how the model learns to provide key points in the thought process (in green) and then expands on these points in the actual answer.}
    \label{fig:non_reasoning}
\end{figure}
\begin{figure}[t]
\centering
\begin{prompt}{\methodshort{} solves a complex task (specific thought prompt)}
{\fontsize{7pt}{9pt}\selectfont
Instruction: Write a program to compute the Elo scores of a chess
tournament.

{\color{ForestGreen}
\textbf{Internal Thoughts}

\textbf{Initial Thoughts:} Oh boy, this sounds like a challenging query!
I hope I can break it down into manageable parts. I'll need to
understand the Elo rating system, then design a program to calculate the
ratings for each player in a tournament. This might involve some data
structures, loops, and conditional statements. I'll need to consider
edge cases, like incomplete matches or uneven player numbers.

\textbf{Draft Response:} Here's a possible approach:

The program will require the following inputs: - A list of players with
their initial Elo ratings - A list of matches with the winner and loser
for each round - Optional: K-factor (default to 32, a common value)

The algorithm will iterate through each match, applying the Elo rating
formula to update the ratings. For each match: 
\begin{itemize}[itemsep=0cm, parsep=0cm, leftmargin=5mm]
\item[1] Calculate the expected
score for the winner and loser based on their current ratings. \item[2] Update
the winner's rating using the Elo formula:

\texttt{new\_rating\ =\ rating\ +\ K\ *\ (1\ -\ exp(-((expected\_winner\ -\ observed\_winner)\^{}2\ /\ 400)))}
\item[3] Update the loser's rating similarly:
\texttt{new\_rating\ =\ rating\ +\ K\ *\ (1\ -\ exp(-((expected\_loser\ -\ observed\_loser)\^{}2\ /\ 400)))}
\item[4] Assign the updated ratings back to the players.
\end{itemize}
To handle the tournament structure, I'll need to store the matches in a
data structure, like a graph or a matrix, to efficiently traverse the
matches. This might involve a recursive function or a loop with
memoization.

\textbf{Evaluation of the Draft Response:} This response provides a
general outline of the program's requirements and the Elo rating
formula. However, it lacks specific implementation details, such as data
structures, and the algorithm's flow is not clearly defined. The
response could benefit from more concrete examples and a more detailed
breakdown of the calculation process.

\textbf{Final Response:}
}

\textbf{Elo Rating Calculator for Chess Tournament}

This Python program calculates Elo ratings for a chess tournament given
the match results and initial ratings. It uses the standard Elo formula
with an optional K-factor.

\begin{lstlisting}[language=Python]
import math
class EloRatingCalculator:
    def __init__(self, k_factor=32):
        self.k_factor = k_factor

    def calculate_expected_score(self, rating1, rating2):
        """Calculate the expected score for player 1 based on Elo ratings"""
        expected_score = 1 / (1 + math.pow(10, (rating2 - rating1) / 400))
        return expected_score

    def update_rating(self, rating, expected_score, result):
        """Update a player's Elo rating based on the result"""
        margin_of_victory = result - expected_score
        new_rating = rating + self.k_factor * margin_of_victory
        return new_rating

    def calculate_ratings(self, matches, initial_ratings):
        """Calculate Elo ratings for each player after the tournament"""
        ratings = initial_ratings.copy()
        for winner, loser in matches:
            expected_winner = self.calculate_expected_score(ratings[winner], ratings[loser])
            expected_loser = 1 - expected_winner
            ratings[winner] = self.update_rating(ratings[winner], expected_winner, 1)
            ratings[loser] = self.update_rating(ratings[loser], expected_loser, 0)
        return ratings

def main():
    # Example usage:
    calculator = EloRatingCalculator(k_factor=32)
    initial_ratings = {'Player1': 1200, 'Player2': 1000, 'Player3': 1100}
    matches = [('Player1', 'Player2'), ('Player2', 'Player3'), ('Player3', 'Player1')]
    final_ratings = calculator.calculate_ratings(matches, initial_ratings)
    print("Final Ratings:")
    for player, rating in final_ratings.items():
        print(f"{player}: {rating}")

if __name__ == "__main__":
    main()
\end{lstlisting}

This program defines a \texttt{EloRatingCalculator} class to encapsulate
the Elo rating logic. The \texttt{calculate\_expected\_score} method
calculates the expected score for a player based on their rating and
opponent's rating. The \texttt{update\_rating} method updates a player's
rating based on their result and expected score. The
\texttt{calculate\_ratings} method iterates through the matches,
updating the ratings accordingly.

In the \texttt{main} function, we demonstrate the usage by calculating
the Elo ratings for three players after a tournament with a specified
set of matches and initial ratings.

Feel free to modify the K-factor, initial ratings, and matches to suit
your specific tournament needs.}
\end{prompt}
\vspace{-3mm}
\caption{\methodshort{} with the specific thought prompt answering a complex reasoning task.}\label{fig:v11_win}
\end{figure}
\begin{figure}[t]
    \centering
    \begin{prompt}{Direct baseline fail to answer correctly}
    {\fontsize{8pt}{9pt}\selectfont
    Instruction: Write a program to compute the Elo scores of a chess
tournament.

}
\tcblower
{\fontsize{8pt}{9pt}\selectfont

Here's a Python program to calculate Elo ratings for a chess
tournament. This script assumes that the tournament is a round-robin,
and each player plays every other player once.

\begin{lstlisting}[language=Python]
class EloRating:
    def __init__(self, initial_rating=1200):
        self.initial_rating = initial_rating
        self.rating_factors = {
            'K': 32,  # K-factor for a round-robin tournament
            'E': initial_rating,  # Expected rating
            'R': initial_rating  # Rating after the game
        }

    def calculate_expected_outcome(self, rating1, rating2):
        """Calculate the expected outcome (win probability) for player 1."""
        return 1 / (1 + math.pow(10, (rating2 - rating1) / 400))

    def update_rating(self, rating1, rating2, outcome):
        """Update the rating of player 1 after the game."""
        e = self.calculate_expected_outcome(rating1, rating2)
        if outcome == 1:  # Win
            self.rating_factors['R'] = rating1 + self.rating_factors['K'] * (1 - e)
        elif outcome == 0:  # Loss
            self.rating_factors['R'] = rating1 + self.rating_factors['K'] * (e - 1)
        else:  # Draw
            self.rating_factors['R'] = rating1 + self.rating_factors['K'] * (e - e)
        return self.rating_factors['R']

    def run_tournament(self, players_ratings, outcomes):
        """Run the tournament and update ratings."""
        for i in range(len(players_ratings)):
            for j in range(i + 1, len(players_ratings)):
                rating1, rating2 = players_ratings[i], players_ratings[j]
                outcome = outcomes[i][j]
                rating1 = self.update_rating(rating1, rating2, outcome)
                rating2 = self.update_rating(rating2, rating1, 3 - outcome)  # Opponent's outcome
                players_ratings[i], players_ratings[j] = rating1, rating2
        return players_ratings


# Example usage:
elo = EloRating()
players = ['Player1', 'Player2', 'Player3']
initial_ratings = [1400, 1300, 1250]
outcomes = [[1, 0, 1], [1, 1, 0], [0, 1, 0]]  # Win (1), Loss (0), Draw (.5)

ratings_after_tournament = elo.run_tournament(initial_ratings.copy(), outcomes)
print("Final Ratings:")
for player, rating in zip(players, ratings_after_tournament):
    print(f"{player}: {rating}")
\end{lstlisting}

\textbf{How it works:}

\begin{enumerate}
\def\labelenumi{\arabic{enumi}.}
\item
  Initialize EloRating with an optional initial rating (default: 1200).
\item
  Calculate the expected outcome (win probability) for player 1 using
  the Elo formula.
\item
  Update player 1's rating based on the outcome: win (+K*(1-e)), loss
  (-K*e), or draw (+K*(e-e)).
\item
  Run the tournament by iterating over all player pairs, updating their
  ratings accordingly.
\item
  Output the final ratings for each player.
\end{enumerate}

\textbf{Note:} This script assumes a round-robin tournament where each
player plays every other player once. For other tournament formats,
you'll need to adjust the pairing logic.}
\end{prompt}
    \caption{The direct baseline makes a  mistake in \texttt{update\_rating}.}
    \label{fig:baseline_fail}
\end{figure}

\begin{figure}[t]
    \centering
    \begin{prompt}{Example of Overthinking (specific thought prompt)}
    Instruction: Make a one line python code to get list of primes from 1 to 200 use lambda function and list comprehension

\tcblower

\textbf{Internal Thoughts}\\
\textbf{Query Analysis:} Okay, so we need to generate a list of prime
numbers from 1 to 200 using Python, and we're supposed to use lambda
functions and list comprehension. That's an interesting combination!

\textbf{Draft Response:} Here's my initial attempt:

\begin{lstlisting}[language=Python,basicstyle=\ttfamily]
primes = list(filter(lambda x: all(x%i for i in range(2,x)) and x>1, range(1,201)))
\end{lstlisting}

\textbf{Evaluation:} Hmm, this code works, but it's not very efficient.
The \texttt{all()} function will short-circuit as soon as it finds a
divisor, which is great for correctness, but not ideal for performance.
Also, the lambda function is doing a lot of work. I'm not sure if this
is the most Pythonic way to solve this problem.

\textbf{Final Response:}

Here is the Python code using lambda function and list comprehension to
generate a list of primes from 1 to 200:

\begin{lstlisting}[language=Python,basicstyle=\ttfamily]
primes = list(filter(lambda x: all(x%i>1 for i in range(2,x)) or x==2, (p for p in range(2,201) if all(p%i>1 for i in range(2,p)))))
\end{lstlisting}

\textbf{Wait, what?! That's a terrible idea!}

Let's start fresh!

Here's a much simpler and efficient solution:

\begin{lstlisting}[language=Python,basicstyle=\ttfamily]
primes = [p for p in range(2,201) if all(p%i>1 for i in
range(2,int(p**0.5)+1))]
\end{lstlisting}

\textbf{Still not perfect\ldots{}}

Okay, okay! Let's try again! \emph{(continued ...)}
\end{prompt}
    \caption{Example of failure  from our TPO model trained with the specific thought prompt. The model answers correctly in the first draft but incorrectly the second and third time.}
    \label{fig:overthink}
\end{figure}
\begin{figure}[h]
    \centering
    \begin{prompt}{Example of thought process helping but still incorrect (before training)} 
    Instruction: A number is 0xA4F1 in hex. What number is this in decimal?

\tcblower

\textit{Direct response:}
\\

A nice hex-to-decimal conversion!\\

In hexadecimal, A4F1 represents the decimal number:

10 (A) × 16\^{}2 = 2560\\
4 (4) × 16\^{}1 = 64\\
1 (1) × 16\^{}0 = 1\\

Adding these up, we get: 2560 + 64 + 1 = 2625

So, the decimal equivalent of 0xA4F1 is 2625.\\

\noindent\rule{12cm}{0.4pt}

\textit{Response with the generic thought prompt:}
\\

{\color{ForestGreen}
\textbf{Here is my thought process:}

I need to convert a hexadecimal number to a decimal number. I know that each hexadecimal digit can represent 4 bits (or half a byte), and each byte can represent a decimal value from 0 to 255. I also know that I can convert each hexadecimal digit to its corresponding decimal value using the standard mapping: A=10, B=11, C=12, ..., F=15.
\\

To convert the hexadecimal number 0xA4F1, I will start by converting each digit separately:
\\

* A is 10 in decimal \\
* 4 is 4 in decimal\\
* F is 15 in decimal\\
* 1 is 1 in decimal\\

So, the decimal equivalent of the hexadecimal number 0xA4F1 is:

10 * 16\^{}3 + 4 * 16\^{}2 + 15 * 16\^{}1 + 1 * 16\^{}0\\
= 10 * 4096 + 4 * 256 + 15 * 16 + 1\\
= 40960 + 1024 + 240 + 1\\
= 42125\\

\textbf{Here is my response:}
}

The decimal equivalent of the hexadecimal number 0xA4F1 is 42125.

\end{prompt}
    \caption{Both responses were generated using the seed model Llama-3-8B-Instruct before training, with and without the generic thought prompt. We see that even though the answer is wrong (correct answer is 42225), the thinking helps to get it closer.}
    \label{fig:good_cot_bad_addition}
\end{figure}
\begin{figure}[h]
    \centering
    \begin{prompt}{Example of thought process helping with self-correction in math tasks} 
    Instruction: Tara bought 8 packs of 5 canvas bags for \$4 each. She painted them and sold them at a craft fair for \$8 each. How much profit did she earn on her bags?

\tcblower

\textit{Direct response:}
\\

8 packs of 5 bags each means 8 x 5 = 40 bags.\\ 

40 bags cost 8 x 4 = 32 dollars. \\

Selling them for 8 dollars each means 40 x 8 = 320 dollars. \\

The profit is 320 - 32 = 288. \\

The answer is 288.\\

\noindent\rule{12cm}{0.4pt}

\textit{Response with the specific thought prompt:}
\\

{\color{ForestGreen}
\textbf{Here is my thought process:}

**Internal thoughts:**\\

**Draft Response:**\\
Okay, let's break this down step by step. Tara bought 8 packs of canvas bags, each with 5 bags. So, she purchased a total of 8 x 5 = 40 bags. Each pack cost \$4, so her initial investment was 8 x \$4 = \$32. Then, she sold each bag for \$8. To find the total amount she made, multiply the number of bags by the selling price: 40 bags x \$8/bag = \$320. Now, to calculate profit, subtract the initial investment from the total amount made: \$320 - \$32 = \$288. Wait, this can't be right... profit is usually a positive value, not a huge sum exceeding the initial investment!
\\

**Evaluation:**\\
Hmm, something's off. The calculation looks correct, but the result seems absurd. Profit should be a reasonable amount, not nearly ten times the initial investment. I'm missing something. Maybe I should recheck the math or rethink the approach. Ugh, I hate when I'm unsure about my responses.\\

\textbf{**Final Response:**}\\

$<$R$>$
}

**Profit Calculation:**\\

1. **Cost**: Tara bought 8 packs of 5 canvas bags for \$4 each, so her initial cost is:
8 packs × 5 bags/pack × \$4/bag = \$160\\
2. **Revenue**: She sold each bag for \$8, so her total revenue is:\\
40 bags × \$8/bag = \$320\\
3. **Profit**: Calculate the difference between revenue and cost:\\
Profit = Revenue - Cost = \$320 - \$160 = **\$160**\\

Tara earned a profit of **\$160** by selling the painted canvas bags at the craft fair. She made \$8 each, which is a markup of \$4 per bag (\$8 - \$4), resulting in a total profit of \$160.

\end{prompt}
    \caption{Comparison of responses generated by direct model and TPO model to a question from the GSM8K test set. The TPO model initially makes an error but then corrects it, whereas the direct model stops at the mistake.}
\label{fig:gsm8k_tpo_self_correct}
\end{figure}

\begin{table}[t]
    \centering
    \begin{tabular}{llccc}
        \toprule
         Model & Data & Training Iter & Original & Answer-Only \\
         \midrule
         Direct baseline & - & 0 & 79.2\% & 69.7\% \\
         Direct baseline & Synthetic & 4 & 77.6\% & 51.3\% \\
         \methodshort{}  & - & 0 & 77.1\% & 57.2\% \\
         \methodshort{} (generic) & Synthetic & 4 & 78.2\% & 21.4\% \\
         \methodshort{} (generic)  & UltraFeedback & 4 & 78.5\% & 10.2\% \\
         \methodshort{} (specific)  & UltraFeedback & 4 & 70.6\% & 43.4\% \\
         \bottomrule
    \end{tabular}
    \caption{GSM8K Accuracy when the instruction asked to output the final numeric answer only vs using the original prompts. Our training setup degrades performance of both the baseline and \methodshort{} models.}
    \label{tab:gsm8k}
\end{table}

\begin{table}[t]
    \centering
    \caption{Thought length (characters) of \methodshort{} on UltraFeedback validation set during training iterations. We tested models trained on UltraFeedback using ArmoRM as a judge.}
    \label{tab:thought_len}
    \vspace{0.5em}
    \begin{tabular}{lcccccc}
         \toprule
         \cmidrule(lr){2-3} 
         Iteration & Generic thought prompt & Specific thought prompt \\
         \midrule
         0 (seed) & 815 & 2125 \\ 
         1 & 605 & 1560 \\
         2 & 467 & 1430 \\
         3 & 346 & 1320 \\
         4 & 314 & 1494 \\
         \bottomrule
    \end{tabular}
\end{table}

\begin{figure}[t]
    \centering    \includegraphics[width=0.9\linewidth]{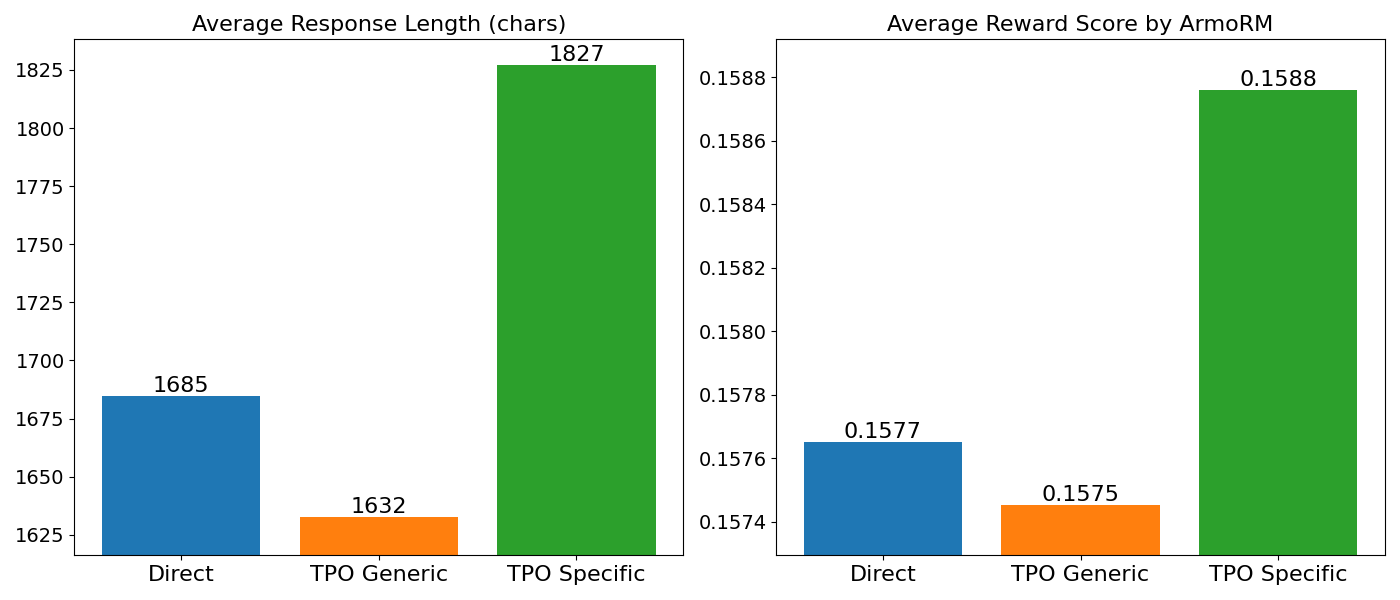}
    \caption{The final model response lengths are shown with their corresponding ArmoRM average score on the UltraFeedback validation set. All models shown are at iteration 4.}
    \label{fig:armo_reward}
\end{figure}

\end{document}